\documentclass{article}

\usepackage{amssymb}
\usepackage{arxiv}
\usepackage{algorithm}
\usepackage{algpseudocode}
\usepackage{xcolor}
\usepackage[utf8]{inputenc} 
\usepackage[T1]{fontenc}    
\usepackage{hyperref}       
\usepackage{url}            
\usepackage{booktabs}       
\usepackage{amsfonts}       
\usepackage{nicefrac}       
\usepackage{microtype}      
\usepackage{lipsum}
\usepackage{graphicx}

\usepackage{pifont}
\usepackage{bbding}
\graphicspath{ {./images/} }

\title{D-RMGPT: Robot-assisted collaborative tasks driven by large multimodal models}

\author{
 Matteo Forlini \\
  Department of Industrial Engineering and Mathematical Sciences\\
  Polytechnic University of Marche\\
  60131 Ancona, Italy \\
  \texttt{m.forlini@pm.univpm.it} \\
   \And
 Mihail Babcinschi \\
  Department of Mechanical Engineering\\
  University of Coimbra\\
  3030-788 Coimbra, Portugal \\
  \texttt{mihail.babcinschi@uc.pt} \\
  \And
 Giacomo Palmieri \\
  Department of Industrial Engineering and Mathematical Sciences\\
  Polytechnic University of Marche\\
  60131 Ancona, Italy \\
  \texttt{g.palmieri@staff.univpm.it} \\
  \And
 Pedro Neto \\
  Department of Mechanical Engineering\\
  University of Coimbra\\
  3030-788 Coimbra, Portugal \\
  \texttt{pedro.neto@dem.uc.pt} \\
}

\begin{document}
\maketitle
\begin{abstract}
Collaborative robots are increasingly popular for assisting humans at work and daily tasks. However, designing and setting up interfaces for human-robot collaboration is challenging, requiring the integration of multiple components, from perception and robot task control to the hardware itself. Frequently, this leads to highly customized solutions that rely on large amounts of costly training data, diverging from the ideal of flexible and general interfaces that empower robots to perceive and adapt to unstructured environments where they can naturally collaborate with humans. To overcome these challenges, this paper presents the Detection-Robot Management GPT (D-RMGPT), a robot-assisted assembly planner based on Large Multimodal Models (LMM). This system can assist inexperienced operators in assembly tasks without requiring any markers or previous training. D-RMGPT is composed of DetGPT-V and R-ManGPT. DetGPT-V, based on GPT-4V(vision), perceives the surrounding environment through one-shot analysis of prompted images of the current assembly stage and the list of components to be assembled. It identifies which components have already been assembled by analysing their features and assembly requirements. R-ManGPT, based on GPT-4, plans the next component to be assembled and generates the robot’s discrete actions to deliver it to the human co-worker. Experimental tests on assembling a toy aircraft demonstrated that D-RMGPT is flexible and intuitive to use, achieving an assembly success rate of 83\% while reducing the assembly time for inexperienced operators by 33\% compared to the manual process.  \url{http://robotics-and-ai.github.io/LMMmodels/} 
\end{abstract}


\section{Introduction}

\begin{figure*}
\centering
\includegraphics[width=1\textwidth]{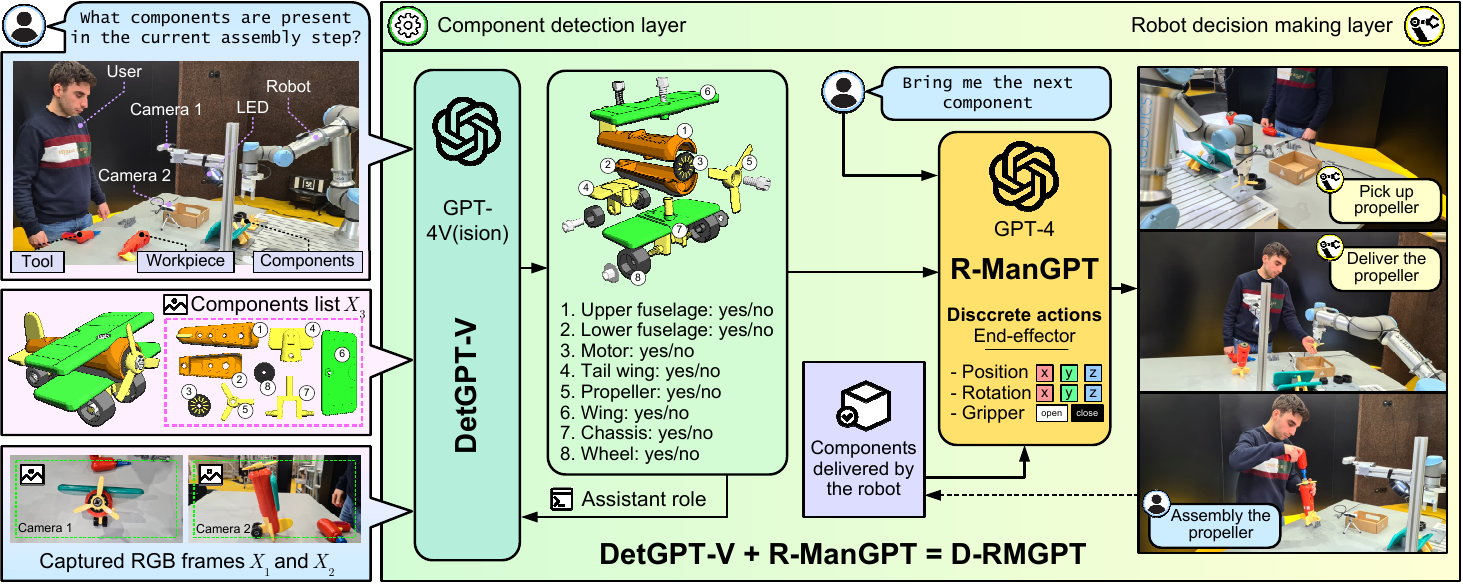}
\caption{The D-RMGPT architecture comprises the detection module DetGPT-V and the robot management and planner module R-ManGPT. These modules, based on GPT-4V(ision) and GPT-4, enable an inexperienced operator to successfully complete an assembly task assisted by a collaborative robot.\label{fig1_main_concept}}
\end{figure*}

Collaborative robots are increasingly present and widespread in a wide variety of application domains, aiming to reduce human effort, structure collaborative work, and enhance productivity. The effectiveness of these robots is highly dependent on the intuitiveness and robustness of the human-robot interfaces and collaborative processes. Significant advancements have been achieved in the field of human-robot interaction (HRI) in recent years. However, these advancements often result in structured applications tailored to specific tasks that are challenging to configure and lack flexibility. Moreover, key elements of a HRI system, such as perception and reasoning, often rely on deep learning or reinforcement learning methodologies that require large amounts of training data. In this context, advances in HRI require general, reliable and integrated solutions to empower robots to perceive the surrounding environment, learning from human co-workers and previous experiences, and reasoning in uncertain conditions. 

Recent advancements in foundation models such as Large Language Models (LLM), Vision Language Models (VLM) and Large Multimodal Models (LMM), have demonstrated significant potential to support robot planning activities and related HRI applications \cite{firoozi2023foundationmodelsroboticsapplications}. While LMM trained in specific application datasets offer superior performance, they can limit the general applicability and flexibility of the robotic system \cite{octomodelteam2024octoopensourcegeneralistrobot}. On the other hand, using off-the-shelf models can strike a good balance between performance and general applicability, without the need for additional training data and model retraining \cite{wake2024gpt4visionroboticsmultimodaltask, zhou2023generalizablelonghorizonmanipulationslarge}. 

Assembly is one of the most common tasks in manufacturing, spanning from fully automated systems to semi-automated to fully manual assembly. The automation of assembly processes can be difficult to achieve, as many tasks require dexterity skills. This scenario opens the door to human-robot collaborative assembly, leveraging the best abilities of both robots and humans for a more productive and less error-prone assembly workflow \cite{WANG2019701}. In addition, each actor can focus on tasks where they perform best, humans can handle tasks requiring high dexterity, while robots can perform monotonous tasks demanding physical effort. Having robots doing such tasks is crucial for companies in the current context of skilled labor shortages. However, the integration of robots in collaborative assembly operations is challenging due to the high variability of the process, the required dexterity, and the need for powerful and reliable perception systems. These are limiting factors for the deployment of collaborative robots.

This study proposes the Detection-Robot Management GPT (D-RMGPT), a robot-assisted assembly planner based on off-the-shelf GPT-4 and GPT-4V(ision). This system is designed to assist inexperienced operators in human-robot collaborative assembly tasks without requiring any markers or prior local training. D-RMGPT comprises two main components: the detection module DetGPT-V and the robot management and planner module R-ManGPT, Fig. \ref{fig1_main_concept}. DetGPT-V perceives the surrounding environment through one-shot analysis of prompted images of the current assembly stage, identifying which components are already assembled. It only requires an image with the full list of components to assemble and their assembly precedence relationships as input, eliminating the need to identify missing components. R-ManGPT acts as operations and robot task manager, planning the next feasible components to assemble and generating the robot’s discrete actions. This work resulted in the following contributions:

\begin{itemize}
    \item D-RMGPT, a novel flexible robot-assisted assembly planner based on GPT-4 and GPT-4V(ision) to assist inexperienced operators in human-robot collaborative assembly tasks, guiding the operators while adapting to their choices. It can adapt to complete an assembly task even when the operator deviates from the suggested assembly sequence.
    \item A detection module, DetGPT-V, perceives the surrounding environment through one-shot analysis of prompted images. DetGPT-V’s component detection abilities compare favourably with commonly used VLM-based object detectors such as ViLD and OWL-ViT.
    \item An action and robot manager module, R-ManGPT, dynamically plans the next components to assemble and generates the robot’s discrete actions. It is easy to finetune for new application domains, being camera and robot independent.
    \item D-RMGPT significantly improves the intuitiveness, flexibility, availability and general applicability of human-robot collaborative assembly while reducing the total assembly time for inexperienced operators.
\end{itemize}

\subsection{Related Work}
In recent years, foundation models have seen widespread usage in different applications, ranging from text recognition and generation to coding, and even extending to the new frontier of image and video classification and generation \cite{bommasani2021opportunities, liu2021pretrainpromptpredictsystematic}. Due to their intrinsic characteristics of perception, reasoning and semantic knowledge \cite{gurnee2023language, petroni2019language}, foundation models such as LLMs \cite{chowdhery2023palm, achiam2023gpt}, VLMs \cite{minderer2022simple} and LMMs \cite{yang2023dawn} are promising intelligent agents for robotics \cite{firoozi2023foundation, xiao2023robot}. They enable robots to understand and reason about the surrounding environment, plan activities, and support decision-making in applications such as HRI and robot navigation. Human-robot interfaces can be easily handled via a simple prompt request, with the potential of making the interactive process more flexible, easier, intuitive, and adaptable to the needs of different users and application scenarios. However, they also present challenges concerning the understanding of real-world environments, dealing with constraints, and perceiving the physical state of a robot. Often, these systems lack important implicit information, leading to inaccurate and insecure actions that result in task failure. Recent studies address these challenges by providing extra awareness tools to the system, such as detectors like Open Vocabulary Object Detection or value functions used as grounding for LLMs \cite{kim2023regionawarepretrainingopenvocabularyobject, zareian2021open, huang2024grounded}. The release of GPT-4 and GPT-4V(ision) makes it possible to combine perception and reasoning for both language and images, creating a synergy that achieves good performance when applied in HRI \cite{yang2023dawn,bai2023qwen, liu2024visual}.   

In computer vision, deep learning methods have demonstrated state-of-the-art performance in object detection and classification \cite{NIPS2012_c399862d, ahmad2022deep, zhao2019object}. However, these methods present a number of challenges related to their reliability and require large and specific datasets for training \cite{geng2023research}. Foundation models such as VLM, trained on massive amounts of general text-image pairs, can help mitigate these limitations by providing text descriptions of images \cite{dosovitskiy2020image, kirillov2023segment}. Contrastive Language-Image Pre-training (CLIP) and A Large-scale ImaGe and Noisy-text embedding (ALIGN) are examples of models that use contrastive learning by computing the similarity between text and image embeddings using textual and visual encoders \cite{radford2021learning,jia2021scaling}. CLIP-Fields learns mapping from spatial locations, raw RGB-D and odometry data, to semantic embedding vectors \cite{shafiullah2022clip}. The model retrieves information from pretrained image models by back-projecting the pixel labels to 3D space and training the output heads to predict semantic labels using an open-vocabulary object detector. The scene representation can then be used as a spatial database for segmentation, instance identification, semantic search over space, and 3D view localization from images. SimVLM is a VLM architecture that uses a transformer encoder to learn image-prefix pairs and a transformer decoder to generate an output text-based sequence \cite{wang2021simvlm}. VLM can identify and detect different components in an image by processing a text prompt that indicates which components need to be investigated in the scene \cite{gu2021open}. Robots can use this information to grasp objects in a scene they are operating in for the first time without specific training. Grounding Languages-Image-Pre-Training (GLIP) unifies object detection and phrase grounding, which involves identifying the correspondence between phrases in a sentence and objects in an image, by passing both a text prompt and images to the system \cite{li2022grounded}. PartSLIP is a zero/few-shot method for 3D point cloud part segmentation by leveraging pretrained image-language model GLIP \cite{liu2023partsliplowshotsegmentation3d}. Open-World Language Vision Transformer (OWL-ViT) is a vision-language model designed to handle open-vocabulary object detection using a transforming architecture \cite{minderer2022simple}. Grounding DINO reviews open-set object detector designs and proposes a tight fusion approach to better fuse cross-modality information \cite{li2022grounded}. MDETR is a framework for object detection that uses a convolutional backbone to extract visual features and the language model RoBERTa to extract text features \cite{kamath2021mdetr,liu2019roberta}. The features of both modalities are projected to a shared embedding space, concatenated, and fed to a transformer encoder-decoder that predicts the bounding boxes of the objects and their grounding.

The characteristics of foundation models make them a key element in the advancement of robotics \cite{xiao2023robotlearningerafoundation,yu2023languagerewardsroboticskill}. The SocraticModel integrates LLMs and VLMs to combine their different commonsense knowledge, exchange information with each other, and capture new multimodal capabilities without requiring finetuning \cite{zeng2022socratic}. This model has been applied in robotic planning and decision-making, using ViLD for object detection in the scene and LLM for task planning \cite{gu2021open,huang2022language}. SayCan combines low-level skills with LLM, enabling the LLM to provide detailed instructions on how to perform an abstract high-level task in a feasible way for a robot \cite{ahn2022can}. LLM ability to reason over natural language, without any additional training, allows them to guide robots in performing manipulation tasks in real world environment \cite{huang2022inner}. LLM have demonstrated potential to act as zero-shot human models for HRI \cite{zhang2023large}. In particular, GPT-3.5 and FLAN-T5 can model high-level human states without fine-tuning, achieving good results in predicting human behavior. Nevertheless, it has also been shown that LLMs are sensitive to prompt design and have reduced performance in tasks involving numerical, physical, or spatial reasoning. Mobile robot motion planning for dinner table arrangement has been addressed using LLM \cite{ding2023task}. Knowing the objects and their dimensions, GPT-3 can provide the relative position of each item on the table to the robot motion planner. Recently, robot navigation has been achieved by combining environment understanding and common-sense reasoning of long-context VLMs with a low-level navigation policy based on topological graphs \cite{chiang2024mobilityvlamultimodalinstruction, huang2023visuallanguagemapsrobot}. LLM text-davinci-003 generates robot action commands to interactively perceive the environment from multimodal sensor feedback \cite{zhao2023chat}. The LLM guides the selection of the correct object to grasp by asking for feedback from the sensors installed on the robot, such as weight, tactile and sound. An interesting study shows that if pre-trained LLM are large enough and prompted appropriately, they can effectively decompose high-level tasks expressed in natural language into mid-level plans without any further training \cite{huang2022language}. Primitive trajectories for robot manipulation can be synthesized by using LLMs to infer affordances and constraints \cite{huang2023voxposer}. Leveraging LLM code-writing capabilities, they can interact with VLM to compose 3D value maps, grounding the knowledge into the observation space of the agent. PaLM-E integrates the LLM PaLM with a vision transformer to transfer knowledge from visual-language domains into embodied reasoning by passing continuous inputs from sensor modalities such as images, neural 3D representations, or states into the language embedding space of a decoder-only language model \cite{chowdhery2023palm, dehghani2023scaling}. This enables the model to reason about text and other modalities jointly. Using GPT-4V(ision), with an image of the current scene and a list of possible sub-tasks as input, a sequence of robot tasks can be planned to reach a given goal \cite{hu2023look}. RoboGPT is a framework in which GPT-3.5 is used to generate demonstrations that are then used to train a robot \cite{jin2024robotgpt}.

\section{Detection-Robot Management GPT (D-RMGPT)}

\subsection{Problem Statement and Assembly Process }
An aircraft toy from the Yale-CMU-Berkeley object and model dataset \cite{7254318} is used as assembly workpiece to assess the performance of the proposed robot-assisted assembly planner. The aircraft is composed of eight different components, plus a power tool, screws, and nuts available to the operator, Fig. \ref{fig1_main_concept}. The complete assembly can be achieved by following a number of different assembly sequences. However, if certain assembly precedence relationships are not respected, it is not possible to successfully complete the assembly of the aircraft.

\begin{figure}
\centering
\includegraphics[width=0.55\textwidth]{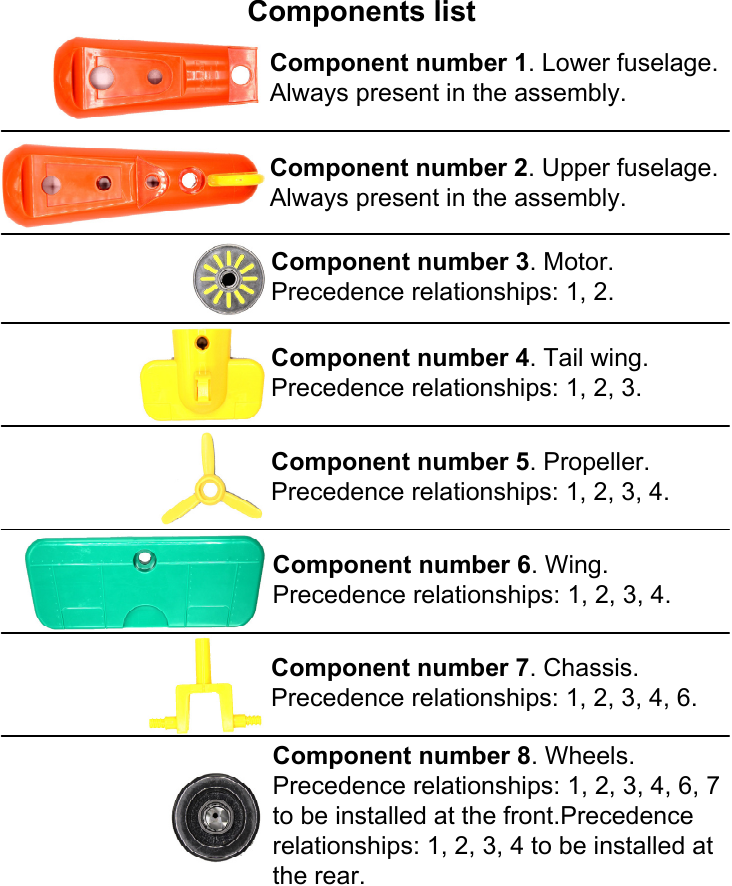}
\caption{Component list image $X_3$. It includes the components picture, number, description and assembly precedence relationships, i.e., the components that need to be assembled before the actual component.   \label{fig:component_list}}
\end{figure}

The assembly process is recommended to start with the two fuselage components, component \#1 and component \#2, Fig. \ref{fig:component_list}. The remaining components are placed in a magazine from which the robot can pick up and deliver them close to the operator, Fig. \ref{fig1_main_concept}. The wheels are divided into a rear pair and a front pair. Since they are identical in color and shape, it is expected that the detection module DetGPT-V will find it challenging to distinguish between them and detect if the wheels are placed at the front or rear of the aircraft toy.

The operator can follow the assembly order defined by the D-RMGPT, or at any step of the assembly, the operator can decide to not follow the D-RMGPT recommendation and choose a different component to assemble on their own initiative. The D-RMGPT should be able to adapt to this unexpected scenario and continue the planning-delivering process to reach the complete assembly.

\subsection{Proposed Approach}
The robot-assisted assembly planner, D-RMGPT, guides the assembly process by suggesting and delivering the necessary components to the human operator in a feasible sequence. At each assembly step, D-RMGPT evaluates the current assembly status, determines which component is needed, and commands the robot to deliver it to the operator. D-RMGPT does not recognize the specific assembly step of the workpiece, instead it identifies the components that are already assembled, making the system less application dependent and consequently more flexible.

The detection module DetGPT-V, based on off-the-shelf GPT-4V, relies on one-shot images from two cameras at each assembly step to assess the current assembly status of the aircraft, identifying which components are already assembled. No textual descriptions of the components or the assembly process sequencing are provided. Only a single image $X_3$ listing all the components and their assembly precedence relationships is required as input at the beginning of the process to fine tune GPT-4V, Fig. \ref{fig:component_list}. The cameras are positioned to capture top and side images of the workbench where the aircraft is assembled, $X_1$ and $X_2$, Fig. \ref{fig1_main_concept}. For this setup, the aircraft workpiece should be placed on the workbench in a position where all components are visible from at least one camera, avoiding occlusions. DetGPT-V provides the list of components present in the current assembly stage, which is saved to a text file and passed to the next iteration through the assistant role. Once the list of components present in that assembly iteration is obtained, it is passed to the robot management and planner module R-ManGPT.

R-ManGPT, based on off-the-shelf GPT-4, plans the next feasible component to assemble and generates the robot’s discrete actions to deliver it to the human operator. Robot actions are defined by the robot end-effector poses (position and orientation) and the gripper state (open or close) for pick-and-place tasks. The components in the magazine are in a known pose for the system, so no markers or additional computer vision systems are required. A general overview of the proposed methodologies and D-RMGPT architecture is provided in Algorithm \ref{algorithm}.

\begin{algorithm}
\caption{D-RMGPT algorithm}\label{algorithm}
\begin{algorithmic}[1]
\Require $X_1, X_2, X_3, \mathcal{L}, det_{t=0} $ \textcolor{cyan}{\Comment{$X_m$ are the input images, $\mathcal{L}$ the prompt request, and $det_{t=0}$ the first components list. }}

\State $t=1$
\State $avail_{t=0}=\{1, ... , n\}$ \textcolor{cyan}{\Comment{List of all the available components at the initial step.}}
\State $brought_{t=0}=\varnothing$ \textcolor{cyan}{\Comment{Components already brought by the robot.}}

\While{$avail_{t-1}\neq \varnothing$}

\State$det_{t}=LMM(X_1, X_2, X_3, \mathcal{L}, det_{t-1})$ \textcolor{cyan}{\Comment{DetGPT-V components identification, where $det_{t}$ is the list of the detected components.}}
\State$bring_{t}=LMM(det_{t}, brought, avail_{t=0})$ \textcolor{cyan}{\Comment{R-ManGPT plans the next component to be picked up and delivered by the robot.}}
\State$brought_t=bring_{t} \cup brought_{t-1}$
\State$avail_{t}=avail_{t=0} - (det_t \cup brought_{t})$
\State$t=t+1$
\EndWhile
\end{algorithmic}
\end{algorithm}

\subsubsection{Prompt Structure}
The components list image $X_3$, Fig. \ref{fig:component_list}, with a size of $768\times2048$ pixels, is analyzed in high detail mode so that GPT-4V does not decrease its resolution by downsampling. It has a cost of 1365 tokens. The current assembly step is captured by the two cameras, which capture top and side images $X_1$ and $X_2$ with a resolution of $680\times480$ pixels, each costing 425 tokens. 

\begin{figure}
\centering
\includegraphics[width=0.64\textwidth]{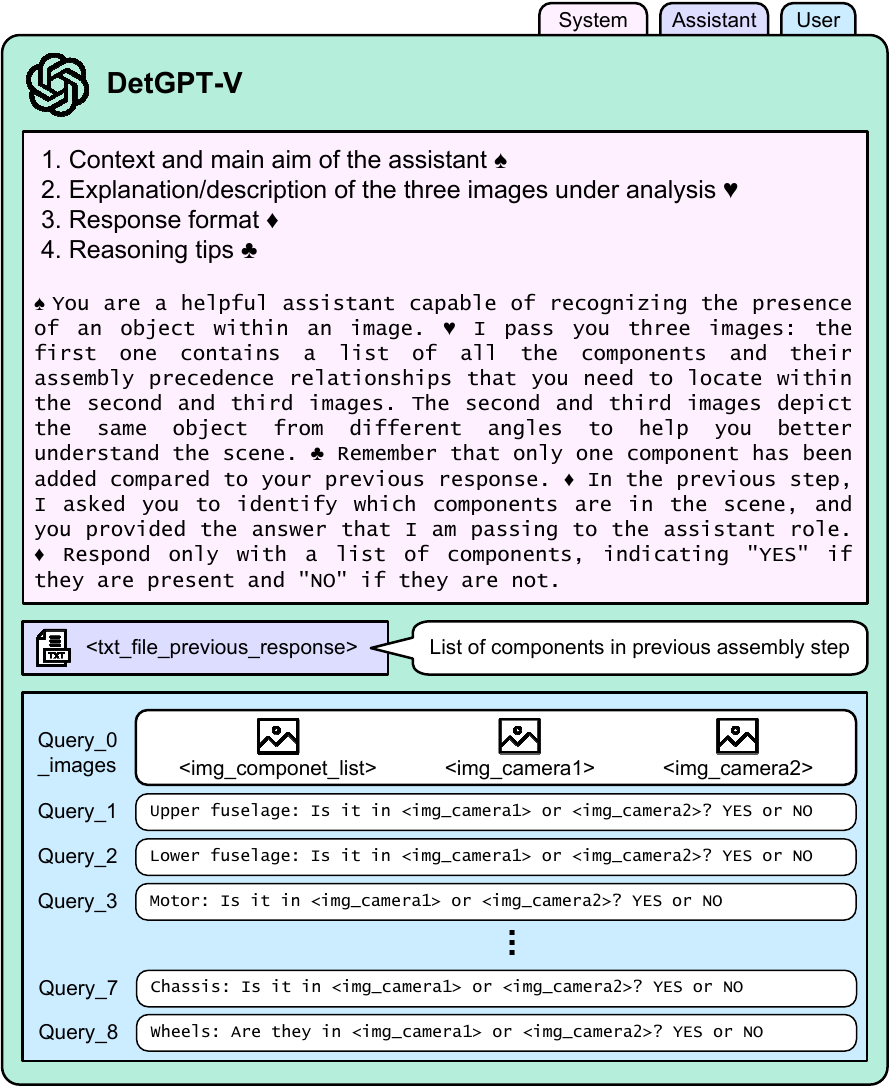}
\caption{Prompt structure for the detection module DetGPT-V.\label{fig: prompt detection}}
\end{figure}

The DetGPT-V prompt structure takes $X_1$, $X_2$ and $X_3$ as input images analyzed by GPT-4V to detect the components already assembled. An example of the response provided by the system is then reported to the assistant role so that the same deterministic output is always given.   In the user role, the three images are loaded, and prompt questions are asked about whether each component is present in the two current state images $X_1$ and $X_2$, requiring a YES or NO answer, Fig. \ref{fig: prompt detection}. To get more accurate and faster responses, a question is asked for each component, indicating YES or NO as possible answers. The output is a list where each component is indicated as present (YES) or not (NO). 
The prompt structure was set up following the instructions and suggestions in \cite{wei2022chain}. Asking to investigate the presence of each individual component in separate queries within the user role allows the identification problem to be decomposed into several subtasks, one for each component, yielding better results in terms of precision and recall. Initial attempts were made by entering all the information in a single initial prompt in the user role, but the response obtained was not deterministic.

\begin{figure}
\centering
\includegraphics[width=0.65\textwidth]{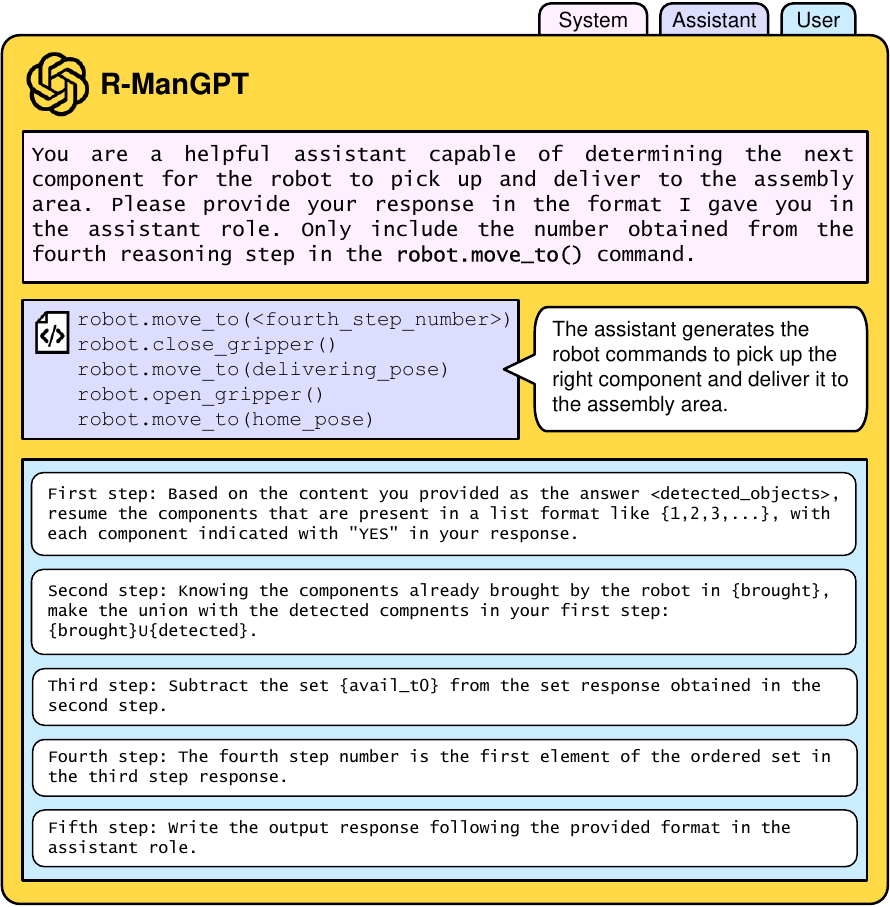}
\caption{Prompt structure for the robot management and planner module R-ManGPT.\label{fig:prompt robot man}}
\end{figure}

The R-ManGPT prompt structure is similar to DetGPT-V, Fig. \ref{fig:prompt robot man}. The global assembly task has been divided into various subtasks. The nouns enclosed in curly brackets represent variables where the information used by the model is stored. The assistant role generates the robot commands for the pick-and-place operation.

\begin{figure*}[h]
\centering
\includegraphics[width=1\textwidth]{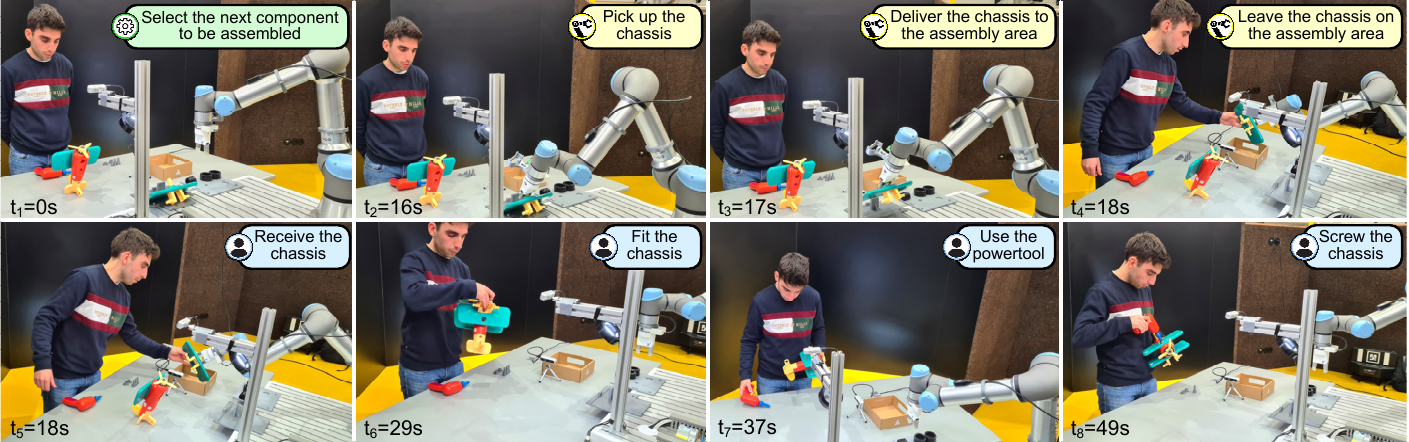}
\caption{Assembly step of the aircraft toy chassis assisted by D-RMGPT.\label{fig:sequences_assembly}}
\end{figure*}

\begin{table*}[h]
\centering
\caption{Experiment \#1 performance results of assembling the aircraft toy assisted by D-RMGPT in a set of 12 tests, each performed by a different inexperienced operator.\label{Tab1}}
\begin{tabular}{cccccc}
\toprule
Test&N° False Positives &N° False Negatives &Total Time [s]&Average GPT Time [s] & Success\\
\midrule
1 & 1$\times$Tail Wing & 0 & 317 & 11 & \textcolor{green}{\Checkmark} \\
2 & 0 & 0 & 345 & 13 & \textcolor{green}{\Checkmark}  \\
3 & 0 & 0 & 303 & 15 & \textcolor{green}{\Checkmark} \\
4 & 0 & 1$\times$Tail Wing & 291 & 16 & \textcolor{green}{\Checkmark} \\
5 & 0 & 0 & 309 & 10 & \textcolor{green}{\Checkmark} \\
6 & 0 & 0 & 263 & 11 & \textcolor{green}{\Checkmark} \\
7 & 0 & 2$\times$Propeller & 289 & 13 & \textcolor{green}{\Checkmark} \\
8 & 0 & 1$\times$Wheel & 319 & 11 & \textcolor{green}{\Checkmark} \\
9 & 0 & 0 & 346 & 19 & \textcolor{green}{\Checkmark} \\
10 & 0 & 2$\times$Tail Wing & 326 & 15 & \textcolor{green}{\Checkmark} \\
11 & 2$\times$Chassis & 2$\times$Tail Wing & - & 19 & \textcolor{red}{\ding{55}} \\
12 & 2$\times$Chassis & 2$\times$Tail Wing & - & 19 & \textcolor{red}{\ding{55}} \\
\bottomrule
\end{tabular}
\end{table*}

\subsection{Experiments and Evaluation}\label{experimental_setup}

\subsubsection{System Setup}

The human operator is working in front of a collaborative robot (5e, Universal Robot, Denmark). The robot picks up and delivers the required aircraft toy components from the magazine to the assembly area in front of the operator, Fig. \ref{fig1_main_concept}. The magazine is also easily accessible to the human, who can choose to take any component manually. The assembly scenes are captured by two cameras (D435i, Intel RealSense, USA), installed to capture top and side views of the assembly area. D-RMGPT communicates with the robot via TCP-IP sockets. 

\subsubsection{Evaluation}

In experiment \#1, the D-RMGPT performance was evaluated in a set of 12 different tests, each conducted by a different inexperienced operator who did not know a priori a feasible assembly sequence for the aircraft toy. The operator’s assembly actions are guided by the proposed D-RMGPT. For each test, we evaluated the false positives and false negatives in the detection phase, the total time required to complete the assembly, the average GPT-4V processing time, and whether the aircraft was successfully assembled.

In experiment \#2, an experienced operator was assisted by D-RMGPT to evaluate the system’s flexibility in a set of 3 tests. At a given assembly step, the operator did not follow the D-RMGPT assembly sequence recommendations, allowing to evaluate how the system reacts and recover from such scenarios.

Finally, in experiment \#3, the assembly process guided by D-RMGPT was performed by 10 different inexperienced operators, without any prior knowledge of the assembly object or previous explanation. Then, another 10 inexperienced operators performed the same assembly process manually, without being guided by D-RMGPT and without robot assistance. The aircraft toy instructions manual for assembly was provided to them 1 minute before starting the assembly. The time taken to complete the assembly was recorded to compare the two testing scenarios.

\subsubsection{Baseline Comparison} \label{baseline_comparison}
The component detection module DetGPT-V was compared with two state-of-the-art VLM-based object detectors, ViLD and OW-ViT. They have been used in popular foundation models robotics applications such as SayCan, SocraticModels and Grounding Detection. Precision and recall are calculated for each model. To make the comparison fair, since GPT-4V is used to identify the presence of the aircraft components but not its position in the image, ViLD and OW-ViT are also used only as detectors. A detected component is considered True Positive if it present in the image and it is identified by ViLD or OW-ViT, even if the bounding box is incorrect (IoU=0\%).

\section{Results and Discussion}

In experiment \#1, for a set of 12 tests, each conducted by a different inexperienced operator, D-RMGPT assisted and guided the operators to complete the assembly of the aircraft toy, Fig. \ref{fig:sequences_assembly}. These operators did not know a feasible assembly sequence beforehand. Table \ref{Tab1} shows the performance values, including false positives and false negatives,  calculated by summing the number of incorrect component detections from all the images analyzed during each test. When a component detection is incorrect, the wrong component is indicated in the Table \ref{Tab1}. The total time required to complete the assembly in the first 10 tests was, on average, approximately 311 seconds. The variability is mainly due to differences in the time required to process information on the OpenAI servers. This is the time D-RMGPT requires at each assembly step to process images and decide the next component to recommend for assembly. The average GPT processing time for each assembly test is shown in Table \ref{Tab1}. Since each assembly involves 7 main calls to OpenAI servers, this processing time accounts for about 30\% of the total assembly time.

\begin{figure*}
\centering
\includegraphics[width=1\textwidth]{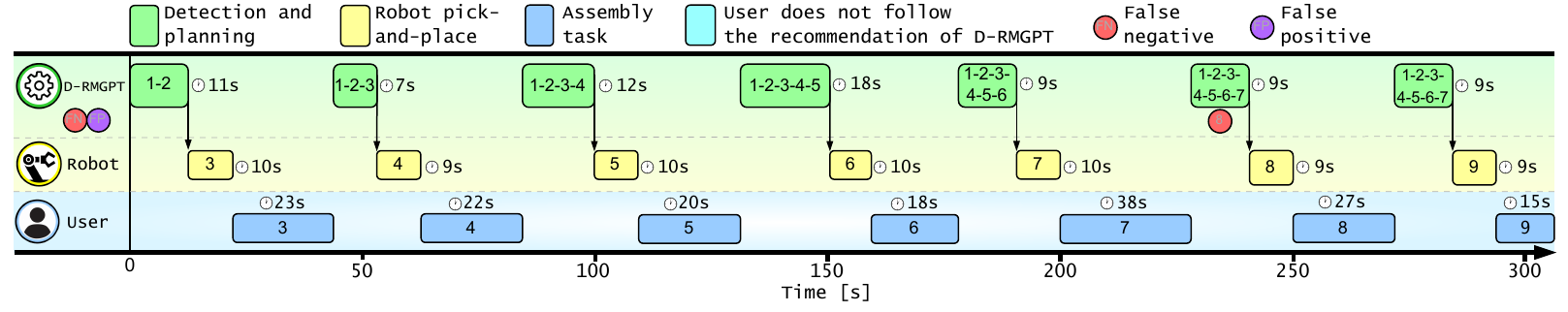}
\caption{Experiment \#1 assembly sequence steps suggested by D-RMGPT and performed by an inexperienced operator. The assembly was successfully achieved despite the presence of false positives and negatives. The numbers inside the boxes represent the component numbers.\label{fig:results_sequence_inexperienced}}
\end{figure*}

\begin{figure*}
\centering
\includegraphics[width=1\textwidth]{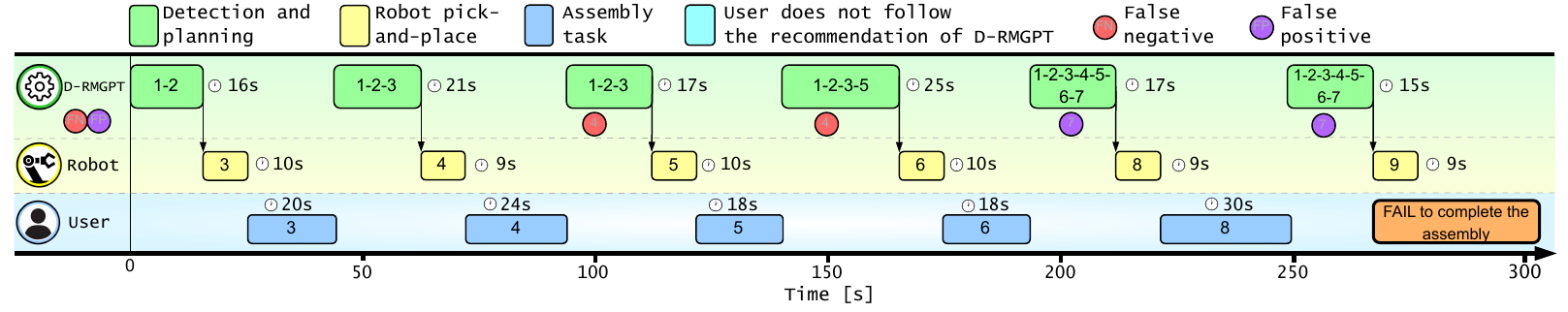}
\caption{Experiment \#1 assembly sequence steps suggested by D-RMGPT, performed by an inexperienced operator, where a false positive detection of the chassis (component \#7) led to the inability to complete the assembly. The numbers inside the boxes represent the component numbers.\label{fig:results_sequence_failure}}
\end{figure*}

\begin{table*}[h]
\centering
\caption{Experiment \#2 results of assembling the aircraft toy guided by D-RMGPT for an experienced operator that did not follow the D-RMGPT recommendations (FP means False Positive, FN means False Negative).\label{Tab2}}
\begin{tabular}{ccccccc}
\toprule
Test&N° FP &N° FN&Total Time [s]&Av. GPT Time [s] & Success & Assembly Order\\
\midrule
1 & 1$\times$Chassis & 1$\times$Motor 1$\times$Propeller 3$\times$Tail Wing  & 252 & 14 & \textcolor{green}{\Checkmark} & 1-2-3-4-8-5-6-7-9\\
2 & 0 & 1$\times$Motor & 273 & 15 & \textcolor{green}{\Checkmark} & 1-2-3-4-6-7-8-5-9 \\
3 & 0 & 0 & 250 & 13 & \textcolor{green}{\Checkmark} & 1-2-3-4-8-6-5-7-9\\
\bottomrule
\end{tabular}
\end{table*}

As shown in Table \ref{Tab1}, out of 12 tests, 10 resulted in successfully completing the assembly task. In the last two tests, the operator was not able to finish the assembly because the system mistakenly recognized the chassis as already installed when it was not. Consequently, the robot never delivered the chassis, which also prevented the installation of subsequent components that required the chassis to be previously assembled. While existing false positives and negatives did not always compromise the assembly, their impact depended on the component. In tests 1, 4, 7 and 8, the system was able to recover from incorrect component detections. Figure \ref{fig:results_sequence_inexperienced} shows an example assembly sequence suggested by D-RMGPT, performed by an inexperienced operator. This example demonstrates that even in the presence of false positives and negatives the system was still able to guide the operator to successfully assemble the aircraft toy. Figure \ref{fig:results_sequence_failure} shows the assembly sequence steps suggested by D-RMGPT in case of a detection failure. A false positive detection of the chassis led to the inability to complete the assembly.

\begin{figure*}
\centering
\includegraphics[width=1\textwidth]{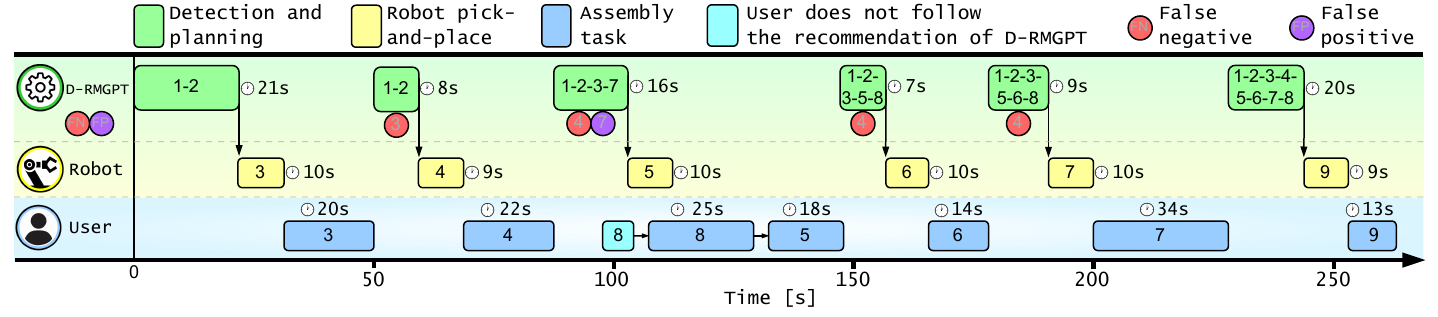}
\caption{Experiment \#2 assembly sequence steps performed by an experienced operator that did not follow the D-RMGPT recommendation in a single assembly step. D-RMGPT managed to successfully complete the assembly. The numbers inside the boxes represent the component numbers.\label{fig:results_sequence_experienced}}
\end{figure*}

In experiment \#2, the D-RMGPT's flexibility in adapting to scenarios where an experienced operator does not follow the assembly recommendations was evaluated. In a set of 3 tests, the operator assembled some of the suggested components while also picking up other components from the magazine, disregarding the suggested ones. The first four components are mandatory to be assembled in order, otherwise, the subsequent components cannot be installed. From the fifth component onward, the order can vary slightly, always respecting the assembly constraints. Table \ref{Tab2} shows the results, demonstrating the system’s ability to adapt and guide the assembly process under these conditions. Figure \ref{fig:results_sequence_experienced} details the assembly steps of test 1, in which after assembling component \#4, the system recommended assembling component \#5. However, the operator did not follow the D-RMGPT recommendation and instead assembled component \#8. In this scenario, D-RMGPT managed to successfully complete the assembly proposing the sequence 1-2-3-4-8-5-6-7-9.

\begin{figure*}
\centering
\includegraphics[width=0.85\textwidth]{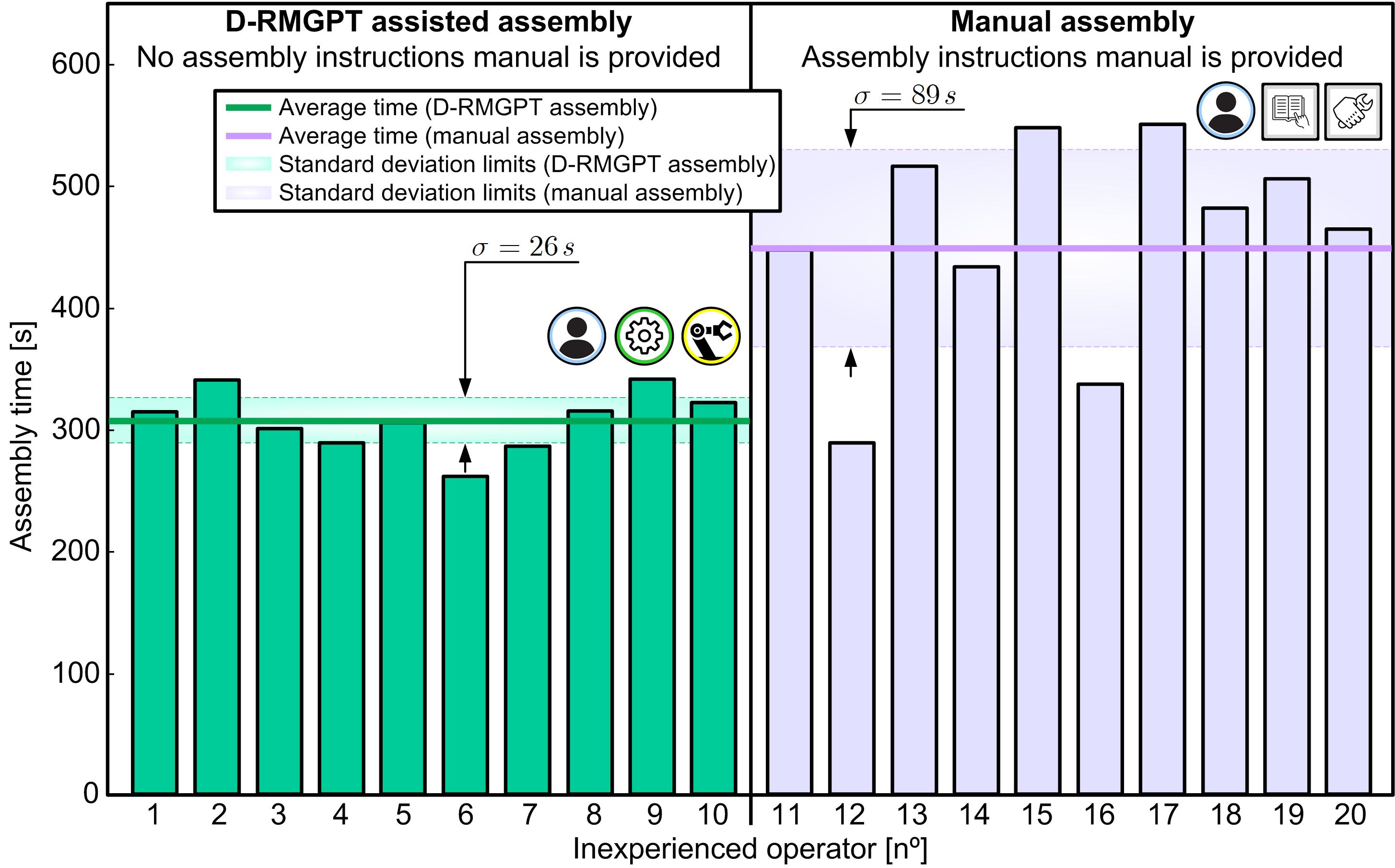}
\caption{Experiment \#3 assembly time of a first group of 10 inexperienced operators assisted by D-RMGPT (green), and a second group of 10 inexperienced operators that performed the assembly manually without any guidance from D-RMGPT or robot assistance (purple). \label{fig:results_time_std}}
\end{figure*}

In experiment \#3, we conducted another set of tests to compare the time required to complete the assembly between two groups of 10 inexperienced operators. The first group was assisted by D-RMGPT, while the second group performed the assembly manually without any guidance from D-RMGPT or robot assistance. The first 10 operators had no prior knowledge of the assembly object or previous explanation about it. For the second group of 10 operators, they performed the assembly manually, without guidance from D-RMGPT or robot assistance, and were provided with the aircraft toy assembly manual of instructions only 1 minute before starting the assembly. The total assembly time for each operator is shown in Fig. \ref{fig:results_time_std}, highlighting the average value and the standard deviation for each group. Results show that the average assembly time for the first group, assisted by D-RMGPT, is $311\,\mathrm{s}$, which is approximately 33\% lower than the average time required for the second group ($467\,\mathrm{s}$). The standard deviation values for the first group ($\sigma=26\,\mathrm{s}$) are smaller than for the second group ($\sigma=89\,\mathrm{s}$), indicating that when the assembly is assisted and guided by D-RMGPT, there are less variability, making the process more uniform and predictable, and less user-dependent. In the second group, some operators assembled components in the wrong sequence and had to disassemble and reassemble them, leading to increased variability in assembly time. In summary, for inexperienced operators, using D-RMGPT reduces both the average assembly time and the variability of the process.

Finally, the component detection module DetGPT-V was compared with state-of-the-art VLM-based object detectors, ViLD and OW-ViT. For each component, the precision $P=\frac{TP}{TP+FP}$  and recall $R=\frac{TP}{TP+FN}$ were calculated in a set of 15 tests. As shown in Table \ref{Tab3}, DetGPT-V offers significantly more accurate detection performance than the other two detectors. For example, in the case of component \#2 (upper fuselage), there is a yellow tail wing on the orange main body, Fig. \ref{fig:component_list}. OWL-ViT and ViLD detectors mistakenly identify this part as a second yellow tail wing component, while DetGPT-V can correctly recognize that it belongs to the upper fuselage component. A common mistake made by these open-vocabulary detectors is that they confuse components of similar shape and color, such as mistaking component \#3 (motor) with component \#8 (wheels). Overall, for this type of application, DetGPT-V, based on GPT-4V, performs better than state-of-the-art VLM-based object detectors like ViLD and OW-ViT.

\begin{table}[h]
\centering
\caption{Component detection baseline comparison between DetGPT-V (based on GPT-4V), ViLD and OW-ViT. Precision (P) and recall (R) are calculated for each model in a set of 15 tests for each component.\label{Tab3}}
\begin{tabular}{lcccccc}
\toprule
Component & \multicolumn{2}{c}{OWL-ViT} & \multicolumn{2}{c}{ViLD} & \multicolumn{2}{c}{DetGPT-V} \\
 & P & R & P & R & P & R \\
\midrule
Fuselage & 1 & 1 & 1 & 1 & 1 & 1\\ 
Motor & 1 & 0.83 & N/A & 0 & 1 & 0.97\\
Tail Wing & 0.80 & 0.80 & 1 & 0.60 & 0.98 & 0.88\\
Propeller & 0.57 & 1 & 0.75 & 0.75 & 1 & 0.94\\
Wing & 1 & 1 & 0.33 & 0.75 & 1 & 1\\
Chassis & 0.50 & 0.50 & 1 & 0.50 & 0.83 & 1\\
Wheel & 0.25 & 1 & 0.50 & 1 & 1 & 0.91\\

\bottomrule
\end{tabular}
\end{table}

\section{Conclusion and Future Work}

This paper introduced D-RMGPT, a robot-assisted assembly planner based on Large Multimodal Models. D-RMGPT demonstrated its capability to guide an assembly process with a robot in the loop. The system achieved an assembly success rate of 83\% while assisting inexperienced operators, without requiring prior information about possible assembly sequences or training data. The results suggest that D-RMGPT is both robust and flexible. It is robust because it can deliver a feasible assembly solution even in the presence of object detection false positives and negatives. It is flexible because it can recover from scenarios where the operator deviates from the assembly recommendations provided by D-RMGPT. This resilience ensures continuity and efficiency in the assembly process, even when unexpected actions are taken by the operator. Additionally, the results demonstrated that D-RMGPT reduces assembly time variability for inexperienced operators, with a 33\% reduction in the assembly time when compared to manual assembly without any guidance from D-RMGPT or robot assistance. The detection module DetGPT-V, a key element of D-RMGPT, compared favorably with state-of-the-art VLM-based object detectors ViLD and OW-ViT in our specific case scenario. Overall, D-RMGPT is a promising and versatile tool, capable of operating in uncertain without requiring specific training data. The system effectively assisted inexperienced operators in assembling a product, demonstrating its potential as a robust and adaptable solution for intuitive human-robot interaction and collaboration.

Future work will focus on studying how the number of input images, which is currently limited to a top and a side view, can affect the detector’s performance while balancing it with the GPT processing time. Additionally, D-RMGPT will be enhanced with the ability to learn operator preferences, further improving the user experience and making the system more intuitive to use.

\section{Acknowledgements}
This work was supported by Portuguese national funds through Fundação para a Ciência e a Tecnologia [grant numbers UIDB/00285/2020, LA/P/0112/2020], and the EuGen program EU4EU. \\

\bibliographystyle{unsrt}

\begin{thebibliography}{1}


\bibitem{achiam2023gpt}
Josh Achiam, Steven Adler, Sandhini Agarwal, Lama Ahmad, Ilge Akkaya,
  Florencia~Leoni Aleman, Diogo Almeida, Janko Altenschmidt, Sam Altman,
  Shyamal Anadkat, et~al.
\newblock Gpt-4 technical report.
\newblock {\em arXiv preprint arXiv:2303.08774}, 2023.

\bibitem{ahmad2022deep}
Hafiz~Mughees Ahmad and Afshin Rahimi.
\newblock Deep learning methods for object detection in smart manufacturing: A
  survey.
\newblock {\em Journal of Manufacturing Systems}, 64:181--196, 2022.

\bibitem{ahn2022can}
Michael Ahn, Anthony Brohan, Noah Brown, Yevgen Chebotar, Omar Cortes, Byron
  David, Chelsea Finn, Chuyuan Fu, Keerthana Gopalakrishnan, Karol Hausman,
  et~al.
\newblock Do as i can, not as i say: Grounding language in robotic affordances.
\newblock {\em arXiv preprint arXiv:2204.01691}, 2022.

\bibitem{bai2023qwen}
Jinze Bai, Shuai Bai, Shusheng Yang, Shijie Wang, Sinan Tan, Peng Wang, Junyang
  Lin, Chang Zhou, and Jingren Zhou.
\newblock Qwen-vl: A frontier large vision-language model with versatile
  abilities.
\newblock {\em arXiv preprint arXiv:2308.12966}, 2023.

\bibitem{bommasani2022opportunitiesrisksfoundationmodels}
Rishi Bommasani, Drew~A. Hudson, Ehsan Adeli, Russ Altman, Simran Arora, Sydney
  von Arx, Michael~S. Bernstein, Jeannette Bohg, Antoine Bosselut, Emma
  Brunskill, Erik Brynjolfsson, Shyamal Buch, Dallas Card, Rodrigo Castellon,
  Niladri Chatterji, Annie Chen, Kathleen Creel, Jared~Quincy Davis, Dora
  Demszky, Chris Donahue, Moussa Doumbouya, Esin Durmus, Stefano Ermon, John
  Etchemendy, Kawin Ethayarajh, Li~Fei-Fei, Chelsea Finn, Trevor Gale, Lauren
  Gillespie, Karan Goel, Noah Goodman, Shelby Grossman, Neel Guha, Tatsunori
  Hashimoto, Peter Henderson, John Hewitt, Daniel~E. Ho, Jenny Hong, Kyle Hsu,
  Jing Huang, Thomas Icard, Saahil Jain, Dan Jurafsky, Pratyusha Kalluri,
  Siddharth Karamcheti, Geoff Keeling, Fereshte Khani, Omar Khattab, Pang~Wei
  Koh, Mark Krass, Ranjay Krishna, Rohith Kuditipudi, Ananya Kumar, Faisal
  Ladhak, Mina Lee, Tony Lee, Jure Leskovec, Isabelle Levent, Xiang~Lisa Li,
  Xuechen Li, Tengyu Ma, Ali Malik, Christopher~D. Manning, Suvir Mirchandani,
  Eric Mitchell, Zanele Munyikwa, Suraj Nair, Avanika Narayan, Deepak
  Narayanan, Ben Newman, Allen Nie, Juan~Carlos Niebles, Hamed Nilforoshan,
  Julian Nyarko, Giray Ogut, Laurel Orr, Isabel Papadimitriou, Joon~Sung Park,
  Chris Piech, Eva Portelance, Christopher Potts, Aditi Raghunathan, Rob Reich,
  Hongyu Ren, Frieda Rong, Yusuf Roohani, Camilo Ruiz, Jack Ryan, Christopher
  Ré, Dorsa Sadigh, Shiori Sagawa, Keshav Santhanam, Andy Shih, Krishnan
  Srinivasan, Alex Tamkin, Rohan Taori, Armin~W. Thomas, Florian Tramèr,
  Rose~E. Wang, William Wang, Bohan Wu, Jiajun Wu, Yuhuai Wu, Sang~Michael Xie,
  Michihiro Yasunaga, Jiaxuan You, Matei Zaharia, Michael Zhang, Tianyi Zhang,
  Xikun Zhang, Yuhui Zhang, Lucia Zheng, Kaitlyn Zhou, and Percy Liang.
\newblock On the opportunities and risks of foundation models, 2022.

\bibitem{bommasani2021opportunities}
Rishi Bommasani, Drew~A Hudson, Ehsan Adeli, Russ Altman, Simran Arora, Sydney
  von Arx, Michael~S Bernstein, Jeannette Bohg, Antoine Bosselut, Emma
  Brunskill, et~al.
\newblock On the opportunities and risks of foundation models.
\newblock {\em arXiv preprint arXiv:2108.07258}, 2021.

\bibitem{calli2017yale}
Berk Calli, Arjun Singh, James Bruce, Aaron Walsman, Kurt Konolige, Siddhartha
  Srinivasa, Pieter Abbeel, and Aaron~M Dollar.
\newblock Yale-cmu-berkeley dataset for robotic manipulation research.
\newblock {\em The International Journal of Robotics Research}, 36(3):261--268,
  2017.

\bibitem{7254318}
Berk Calli, Aaron Walsman, Arjun Singh, Siddhartha Srinivasa, Pieter Abbeel,
  and Aaron~M. Dollar.
\newblock Benchmarking in manipulation research: Using the yale-cmu-berkeley
  object and model set.
\newblock {\em IEEE Robotics and Automation Magazine}, 22(3):36--52, 2015.

\bibitem{chiang2024mobilityvlamultimodalinstruction}
Hao-Tien~Lewis Chiang, Zhuo Xu, Zipeng Fu, Mithun~George Jacob, Tingnan Zhang,
  Tsang-Wei~Edward Lee, Wenhao Yu, Connor Schenck, David Rendleman, Dhruv Shah,
  Fei Xia, Jasmine Hsu, Jonathan Hoech, Pete Florence, Sean Kirmani, Sumeet
  Singh, Vikas Sindhwani, Carolina Parada, Chelsea Finn, Peng Xu, Sergey
  Levine, and Jie Tan.
\newblock Mobility vla: Multimodal instruction navigation with long-context
  vlms and topological graphs, 2024.

\bibitem{chowdhery2023palm}
Aakanksha Chowdhery, Sharan Narang, Jacob Devlin, Maarten Bosma, Gaurav Mishra,
  Adam Roberts, Paul Barham, Hyung~Won Chung, Charles Sutton, Sebastian
  Gehrmann, et~al.
\newblock Palm: Scaling language modeling with pathways.
\newblock {\em Journal of Machine Learning Research}, 24(240):1--113, 2023.

\bibitem{dehghani2023scaling}
Mostafa Dehghani, Josip Djolonga, Basil Mustafa, Piotr Padlewski, Jonathan
  Heek, Justin Gilmer, Andreas~Peter Steiner, Mathilde Caron, Robert Geirhos,
  Ibrahim Alabdulmohsin, et~al.
\newblock Scaling vision transformers to 22 billion parameters.
\newblock In {\em International Conference on Machine Learning}, pages
  7480--7512. PMLR, 2023.

\bibitem{ding2023task}
Yan Ding, Xiaohan Zhang, Chris Paxton, and Shiqi Zhang.
\newblock Task and motion planning with large language models for object
  rearrangement.
\newblock {\em arXiv preprint arXiv:2303.06247}, 2023.

\bibitem{dosovitskiy2020image}
Alexey Dosovitskiy, Lucas Beyer, Alexander Kolesnikov, Dirk Weissenborn,
  Xiaohua Zhai, Thomas Unterthiner, Mostafa Dehghani, Matthias Minderer, Georg
  Heigold, Sylvain Gelly, et~al.
\newblock An image is worth 16x16 words: Transformers for image recognition at
  scale.
\newblock {\em arXiv preprint arXiv:2010.11929}, 2020.

\bibitem{firoozi2023foundation}
Roya Firoozi, Johnathan Tucker, Stephen Tian, Anirudha Majumdar, Jiankai Sun,
  Weiyu Liu, Yuke Zhu, Shuran Song, Ashish Kapoor, Karol Hausman, et~al.
\newblock Foundation models in robotics: Applications, challenges, and the
  future.
\newblock {\em arXiv preprint arXiv:2312.07843}, 2023.

\bibitem{firoozi2023foundationmodelsroboticsapplications}
Roya Firoozi, Johnathan Tucker, Stephen Tian, Anirudha Majumdar, Jiankai Sun,
  Weiyu Liu, Yuke Zhu, Shuran Song, Ashish Kapoor, Karol Hausman, Brian Ichter,
  Danny Driess, Jiajun Wu, Cewu Lu, and Mac Schwager.
\newblock Foundation models in robotics: Applications, challenges, and the
  future, 2023.

\bibitem{geng2023research}
Kaiguo Geng, Jinwei Qiao, Na~Liu, Zhi Yang, Rongmin Zhang, and Huiling Li.
\newblock Research on real-time detection of stacked objects based on deep
  learning.
\newblock {\em Journal of Intelligent \& Robotic Systems}, 109(4):82, 2023.

\bibitem{gu2021open}
Xiuye Gu, Tsung-Yi Lin, Weicheng Kuo, and Yin Cui.
\newblock Open-vocabulary object detection via vision and language knowledge
  distillation.
\newblock {\em arXiv preprint arXiv:2104.13921}, 2021.

\bibitem{gurnee2023language}
Wes Gurnee and Max Tegmark.
\newblock Language models represent space and time.
\newblock {\em arXiv preprint arXiv:2310.02207}, 2023.

\bibitem{hu2023look}
Yingdong Hu, Fanqi Lin, Tong Zhang, Li~Yi, and Yang Gao.
\newblock Look before you leap: Unveiling the power of gpt-4v in robotic
  vision-language planning.
\newblock {\em arXiv preprint arXiv:2311.17842}, 2023.

\bibitem{huang2023visuallanguagemapsrobot}
Chenguang Huang, Oier Mees, Andy Zeng, and Wolfram Burgard.
\newblock Visual language maps for robot navigation, 2023.

\bibitem{huang2022language}
Wenlong Huang, Pieter Abbeel, Deepak Pathak, and Igor Mordatch.
\newblock Language models as zero-shot planners: Extracting actionable
  knowledge for embodied agents.
\newblock In {\em International Conference on Machine Learning}, pages
  9118--9147. PMLR, 2022.

\bibitem{huang2023voxposer}
Wenlong Huang, Chen Wang, Ruohan Zhang, Yunzhu Li, Jiajun Wu, and Li~Fei-Fei.
\newblock Voxposer: Composable 3d value maps for robotic manipulation with
  language models.
\newblock {\em arXiv preprint arXiv:2307.05973}, 2023.

\bibitem{huang2024grounded}
Wenlong Huang, Fei Xia, Dhruv Shah, Danny Driess, Andy Zeng, Yao Lu, Pete
  Florence, Igor Mordatch, Sergey Levine, Karol Hausman, et~al.
\newblock Grounded decoding: Guiding text generation with grounded models for
  embodied agents.
\newblock {\em Advances in Neural Information Processing Systems}, 36, 2024.

\bibitem{huang2022inner}
Wenlong Huang, Fei Xia, Ted Xiao, Harris Chan, Jacky Liang, Pete Florence, Andy
  Zeng, Jonathan Tompson, Igor Mordatch, Yevgen Chebotar, et~al.
\newblock Inner monologue: Embodied reasoning through planning with language
  models.
\newblock {\em arXiv preprint arXiv:2207.05608}, 2022.

\bibitem{jia2021scaling}
Chao Jia, Yinfei Yang, Ye~Xia, Yi-Ting Chen, Zarana Parekh, Hieu Pham, Quoc Le,
  Yun-Hsuan Sung, Zhen Li, and Tom Duerig.
\newblock Scaling up visual and vision-language representation learning with
  noisy text supervision.
\newblock In {\em International conference on machine learning}, pages
  4904--4916. PMLR, 2021.

\bibitem{jiang2020can}
Zhengbao Jiang, Frank~F Xu, Jun Araki, and Graham Neubig.
\newblock How can we know what language models know?
\newblock {\em Transactions of the Association for Computational Linguistics},
  8:423--438, 2020.

\bibitem{jin2024robotgpt}
Yixiang Jin, Dingzhe Li, A~Yong, Jun Shi, Peng Hao, Fuchun Sun, Jianwei Zhang,
  and Bin Fang.
\newblock Robotgpt: Robot manipulation learning from chatgpt.
\newblock {\em IEEE Robotics and Automation Letters}, 2024.

\bibitem{kamath2021mdetr}
Aishwarya Kamath, Mannat Singh, Yann LeCun, Gabriel Synnaeve, Ishan Misra, and
  Nicolas Carion.
\newblock Mdetr-modulated detection for end-to-end multi-modal understanding.
\newblock In {\em Proceedings of the IEEE/CVF International Conference on
  Computer Vision}, pages 1780--1790, 2021.

\bibitem{kim2023regionawarepretrainingopenvocabularyobject}
Dahun Kim, Anelia Angelova, and Weicheng Kuo.
\newblock Region-aware pretraining for open-vocabulary object detection with
  vision transformers, 2023.

\bibitem{kirillov2023segment}
Alexander Kirillov, Eric Mintun, Nikhila Ravi, Hanzi Mao, Chloe Rolland, Laura
  Gustafson, Tete Xiao, Spencer Whitehead, Alexander~C Berg, Wan-Yen Lo, et~al.
\newblock Segment anything.
\newblock {\em arXiv preprint arXiv:2304.02643}, 2023.

\bibitem{NIPS2012_c399862d}
Alex Krizhevsky, Ilya Sutskever, and Geoffrey~E Hinton.
\newblock Imagenet classification with deep convolutional neural networks.
\newblock In F.~Pereira, C.J. Burges, L.~Bottou, and K.Q. Weinberger, editors,
  {\em Advances in Neural Information Processing Systems}, volume~25. Curran
  Associates, Inc., 2012.

\bibitem{li2022grounded}
Liunian~Harold Li, Pengchuan Zhang, Haotian Zhang, Jianwei Yang, Chunyuan Li,
  Yiwu Zhong, Lijuan Wang, Lu~Yuan, Lei Zhang, Jenq-Neng Hwang, et~al.
\newblock Grounded language-image pre-training.
\newblock In {\em Proceedings of the IEEE/CVF Conference on Computer Vision and
  Pattern Recognition}, pages 10965--10975, 2022.

\bibitem{liu2024visual}
Haotian Liu, Chunyuan Li, Qingyang Wu, and Yong~Jae Lee.
\newblock Visual instruction tuning.
\newblock {\em Advances in neural information processing systems}, 36, 2024.

\bibitem{liu2023partsliplowshotsegmentation3d}
Minghua Liu, Yinhao Zhu, Hong Cai, Shizhong Han, Zhan Ling, Fatih Porikli, and
  Hao Su.
\newblock Partslip: Low-shot part segmentation for 3d point clouds via
  pretrained image-language models, 2023.

\bibitem{liu2021pretrainpromptpredictsystematic}
Pengfei Liu, Weizhe Yuan, Jinlan Fu, Zhengbao Jiang, Hiroaki Hayashi, and
  Graham Neubig.
\newblock Pre-train, prompt, and predict: A systematic survey of prompting
  methods in natural language processing, 2021.

\bibitem{liu2023pre}
Pengfei Liu, Weizhe Yuan, Jinlan Fu, Zhengbao Jiang, Hiroaki Hayashi, and
  Graham Neubig.
\newblock Pre-train, prompt, and predict: A systematic survey of prompting
  methods in natural language processing.
\newblock {\em ACM Computing Surveys}, 55(9):1--35, 2023.

\bibitem{liu2023grounding}
Shilong Liu, Zhaoyang Zeng, Tianhe Ren, Feng Li, Hao Zhang, Jie Yang, Chunyuan
  Li, Jianwei Yang, Hang Su, Jun Zhu, and Lei Zhang.
\newblock Grounding dino: Marrying dino with grounded pre-training for open-set
  object detection, 2023.

\bibitem{liu2019roberta}
Yinhan Liu, Myle Ott, Naman Goyal, Jingfei Du, Mandar Joshi, Danqi Chen, Omer
  Levy, Mike Lewis, Luke Zettlemoyer, and Veselin Stoyanov.
\newblock Roberta: A robustly optimized bert pretraining approach.
\newblock {\em arXiv preprint arXiv:1907.11692}, 2019.

\bibitem{minderer2022simple}
Matthias Minderer, Alexey Gritsenko, Austin Stone, Maxim Neumann, Dirk
  Weissenborn, Alexey Dosovitskiy, Aravindh Mahendran, Anurag Arnab, Mostafa
  Dehghani, Zhuoran Shen, et~al.
\newblock Simple open-vocabulary object detection.
\newblock In {\em European Conference on Computer Vision}, pages 728--755.
  Springer, 2022.

\bibitem{openAIgpt4v}
OpenAI.
\newblock Gpt-4v(ision) system card, 2023.

\bibitem{blogpromptgpt}
OpenAI.
\newblock Best practices for prompt engineering with the openai api, 2024.

\bibitem{petroni2019language}
Fabio Petroni, Tim Rockt{\"a}schel, Patrick Lewis, Anton Bakhtin, Yuxiang Wu,
  Alexander~H Miller, and Sebastian Riedel.
\newblock Language models as knowledge bases?
\newblock {\em arXiv preprint arXiv:1909.01066}, 2019.

\bibitem{radford2021learning}
Alec Radford, Jong~Wook Kim, Chris Hallacy, Aditya Ramesh, Gabriel Goh,
  Sandhini Agarwal, Girish Sastry, Amanda Askell, Pamela Mishkin, Jack Clark,
  et~al.
\newblock Learning transferable visual models from natural language
  supervision.
\newblock In {\em International conference on machine learning}, pages
  8748--8763. PMLR, 2021.

\bibitem{shafiullah2022clip}
Nur Muhammad~Mahi Shafiullah, Chris Paxton, Lerrel Pinto, Soumith Chintala, and
  Arthur Szlam.
\newblock Clip-fields: Weakly supervised semantic fields for robotic memory.
\newblock {\em arXiv preprint arXiv:2210.05663}, 2022.

\bibitem{octomodelteam2024octoopensourcegeneralistrobot}
Octo~Model Team, Dibya Ghosh, Homer Walke, Karl Pertsch, Kevin Black, Oier
  Mees, Sudeep Dasari, Joey Hejna, Tobias Kreiman, Charles Xu, Jianlan Luo,
  You~Liang Tan, Lawrence~Yunliang Chen, Pannag Sanketi, Quan Vuong, Ted Xiao,
  Dorsa Sadigh, Chelsea Finn, and Sergey Levine.
\newblock Octo: An open-source generalist robot policy, 2024.

\bibitem{wake2024gpt4visionroboticsmultimodaltask}
Naoki Wake, Atsushi Kanehira, Kazuhiro Sasabuchi, Jun Takamatsu, and Katsushi
  Ikeuchi.
\newblock Gpt-4v(ision) for robotics: Multimodal task planning from human
  demonstration, 2024.

\bibitem{wang2018deep}
Jinjiang Wang, Yulin Ma, Laibin Zhang, Robert~X Gao, and Dazhong Wu.
\newblock Deep learning for smart manufacturing: Methods and applications.
\newblock {\em Journal of manufacturing systems}, 48:144--156, 2018.

\bibitem{WANG2019701}
L.~Wang, R.~Gao, J.~Váncza, J.~Krüger, X.V. Wang, S.~Makris, and
  G.~Chryssolouris.
\newblock Symbiotic human-robot collaborative assembly.
\newblock {\em CIRP Annals}, 68(2):701--726, 2019.

\bibitem{wang2021simvlm}
Zirui Wang, Jiahui Yu, Adams~Wei Yu, Zihang Dai, Yulia Tsvetkov, and Yuan Cao.
\newblock Simvlm: Simple visual language model pretraining with weak
  supervision.
\newblock {\em arXiv preprint arXiv:2108.10904}, 2021.

\bibitem{wei2022chain}
Jason Wei, Xuezhi Wang, Dale Schuurmans, Maarten Bosma, Fei Xia, Ed~Chi, Quoc~V
  Le, Denny Zhou, et~al.
\newblock Chain-of-thought prompting elicits reasoning in large language
  models.
\newblock {\em Advances in neural information processing systems},
  35:24824--24837, 2022.

\bibitem{xiao2023robotlearningerafoundation}
Xuan Xiao, Jiahang Liu, Zhipeng Wang, Yanmin Zhou, Yong Qi, Qian Cheng, Bin He,
  and Shuo Jiang.
\newblock Robot learning in the era of foundation models: A survey, 2023.

\bibitem{xiao2023robot}
Xuan Xiao, Jiahang Liu, Zhipeng Wang, Yanmin Zhou, Yong Qi, Qian Cheng, Bin He,
  and Shuo Jiang.
\newblock Robot learning in the era of foundation models: A survey.
\newblock {\em arXiv preprint arXiv:2311.14379}, 2023.

\bibitem{yang2023dawn}
Zhengyuan Yang, Linjie Li, Kevin Lin, Jianfeng Wang, Chung-Ching Lin, Zicheng
  Liu, and Lijuan Wang.
\newblock The dawn of lmms: Preliminary explorations with gpt-4v (ision).
\newblock {\em arXiv preprint arXiv:2309.17421}, 9(1):1, 2023.

\bibitem{yu2023languagerewardsroboticskill}
Wenhao Yu, Nimrod Gileadi, Chuyuan Fu, Sean Kirmani, Kuang-Huei Lee,
  Montse~Gonzalez Arenas, Hao-Tien~Lewis Chiang, Tom Erez, Leonard Hasenclever,
  Jan Humplik, Brian Ichter, Ted Xiao, Peng Xu, Andy Zeng, Tingnan Zhang,
  Nicolas Heess, Dorsa Sadigh, Jie Tan, Yuval Tassa, and Fei Xia.
\newblock Language to rewards for robotic skill synthesis, 2023.

\bibitem{zareian2021open}
Alireza Zareian, Kevin~Dela Rosa, Derek~Hao Hu, and Shih-Fu Chang.
\newblock Open-vocabulary object detection using captions.
\newblock In {\em Proceedings of the IEEE/CVF Conference on Computer Vision and
  Pattern Recognition}, pages 14393--14402, 2021.

\bibitem{zeng2022socratic}
Andy Zeng, Maria Attarian, Brian Ichter, Krzysztof Choromanski, Adrian Wong,
  Stefan Welker, Federico Tombari, Aveek Purohit, Michael Ryoo, Vikas
  Sindhwani, et~al.
\newblock Socratic models: Composing zero-shot multimodal reasoning with
  language.
\newblock {\em arXiv preprint arXiv:2204.00598}, 2022.

\bibitem{zhang2023large}
Bowen Zhang and Harold Soh.
\newblock Large language models as zero-shot human models for human-robot
  interaction.
\newblock {\em arXiv preprint arXiv:2303.03548}, 2023.

\bibitem{zhao2023chat}
Xufeng Zhao, Mengdi Li, Cornelius Weber, Muhammad~Burhan Hafez, and Stefan
  Wermter.
\newblock Chat with the environment: Interactive multimodal perception using
  large language models.
\newblock {\em arXiv preprint arXiv:2303.08268}, 2023.

\bibitem{zhao2019object}
Zhong-Qiu Zhao, Peng Zheng, Shou-tao Xu, and Xindong Wu.
\newblock Object detection with deep learning: A review.
\newblock {\em IEEE transactions on neural networks and learning systems},
  30(11):3212--3232, 2019.

\bibitem{zhou2023generalizablelonghorizonmanipulationslarge}
Haoyu Zhou, Mingyu Ding, Weikun Peng, Masayoshi Tomizuka, Lin Shao, and Chuang
  Gan.
\newblock Generalizable long-horizon manipulations with large language models,
  2023.



\end{thebibliography}

\end{document}